\def\year{2020}\relax
\documentclass[letterpaper]{article} 
\usepackage{aaai20}  
\usepackage{times}  
\usepackage{helvet} 
\usepackage{courier}  
\usepackage[hyphens]{url}  
\usepackage{graphicx} 
\usepackage{graphicx}  
\usepackage{latexsym}
\usepackage{amsmath}
\usepackage{multirow}
\usepackage{booktabs}
\usepackage{subcaption}
\usepackage[ruled, linesnumbered]{algorithm2e}
\usepackage{comment}

\newcommand{\newcite}[1]{\citeauthor{#1}~(\citeyear{#1})}

\title{Neural Snowball for Few-Shot Relation Learning}
\author{
Tianyu Gao\textsuperscript{\rm 1}, 
Xu Han\textsuperscript{\rm 1}, 
Ruobing Xie\textsuperscript{\rm 2}, 
Zhiyuan Liu\textsuperscript{\rm 1}\thanks{\quad Corresponding author: Z.Liu(liuzy@tsinghua.edu.cn)},\\
{\bf \Large
Fen Lin\textsuperscript{\rm 2}, 
Leyu Lin\textsuperscript{\rm 2}, 
Maosong Sun\textsuperscript{\rm 1}
}\\
\textsuperscript{\rm 1}Department of Computer Science and Technology, Tsinghua University, Beijing, China\\
\textsuperscript{\rm 2}Search Product Center, WeChat Search Application Department, Tencent, China\\
\{gty16,hanxu17\}@mails.tsinghua.edu.cn, ruobingxie@tencent.com, liuzy@tsinghua.edu.cn
}
\begin{document}

\maketitle

\begin{abstract}
Knowledge graphs typically undergo open-ended growth of new relations. This cannot be well handled by relation extraction that focuses on pre-defined relations with sufficient training data. To address new relations with few-shot instances, we propose a novel bootstrapping approach, Neural Snowball, to learn new relations by transferring semantic knowledge about existing relations. More specifically, we use Relational Siamese Networks (RSN) to learn the metric of relational similarities between instances based on existing relations and their labeled data. Afterwards, given a new relation and its few-shot instances, we use RSN to accumulate reliable instances from unlabeled corpora; these instances are used to train a relation classifier, which can further identify new facts of the new relation. The process is conducted iteratively like a snowball. Experiments show that our model can gather high-quality instances for better few-shot relation learning and achieves significant improvement compared to baselines. Codes and datasets are released on \url{https://github.com/thunlp/Neural-Snowball}.
\end{abstract}


\section{Introduction}



Knowledge graphs (KGs) such as WordNet~\cite{miller1995wordnet}, Freebase~\cite{bollacker2008freebase} and Wikidata~\cite{vrandevcic2014wikidata} have multiple applications in information retrieval, question answering and recommender systems. Such KGs consist of relation facts with triplet format $(e_h, r, e_t)$ representing a relation $r$ between entities $e_h$ and $e_t$.
Though existing KGs have acquired large amounts of facts, they still have huge growth space compared to real-world data. 
To enrich KGs, relation extraction (RE) is investigated to extract relation facts from plain text. 

One challenge of RE is that novel relations
emerge rapidly in KGs, yet most RE models cannot handle those new relations well since they rely on RE datasets with only
a limited number of predefined relations. 
One of the largest RE dataset, FewRel \cite{han2018fewrel}, only has 100 relations, yet there were already 920 relations in Wikidata in 2014 \cite{vrandevcic2014wikidata}, let alone it contains nearly 6,000 relations now.


\begin{figure}[t]
\centering
\includegraphics[width=0.45\textwidth]{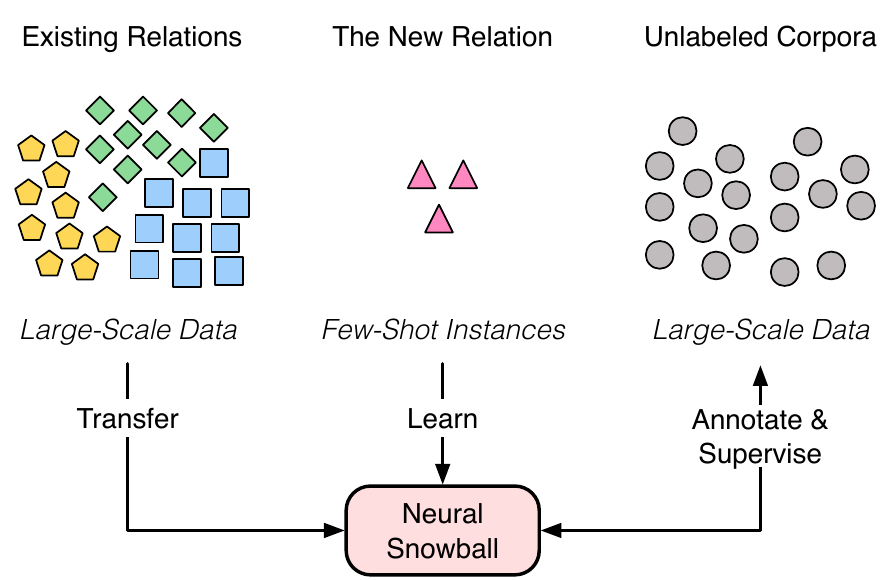}
\caption{An illustration of how Neural Snowball utilizes three different kinds of data to learn new relations.}
\label{fig:ad}
\end{figure}

To extract relation facts of novel relations, many existing approaches have studied bootstrapping RE, which extracts triplets for a new relation with few seed relation facts.
\newcite{brin1998extracting} proposes to extract author-book facts with a small set of (author, book) pairs as input. It iteratively finds mentions of seed pairs from the web, and then extracts sentence patterns from those mentions and finds new pairs by pattern matching. \newcite{agichtein2000snowball} further improve this method and name it as Snowball, for that relation facts and their mentions accumulate like a snowball.



However, most existing bootstrapping models confine themselves to only utilize seed relation facts and fail to take advantage of available large-scale labeled datasets, which have been proved to be a valuable resource. Though data of existing relations might have a very different distribution with new relations, it still can be used to train a deep learning model that extracts abstract features at the higher levels of the representation, suiting both historical and unseen relations \cite{bengio2012deep}. This technique, named as transfer learning, has been widely adopted in image few-shot tasks. Previous work has investigated transferring metrics \cite{koch2015siamese} to measure similarities between objects and meta-information \cite{ravi2016optimization} to fast adapt to new tasks.


Based on bootstrapping and transfer learning, we present \textbf{Neural Snowball} for learning to classify new relations with insufficient training data. 
Given seed instances with relation facts of a new relation, Neural Snowball finds reliable mentions of these facts.
Then they are used to train a relation classifier, which aims at discovering reliable instances with new relation facts. These instances then serve as the inputs of the new iteration.

We also apply Relational Siamese Networks (RSN) to select high-confidence new instances.
Siamese networks \cite{bromley1994signature} usually contain dual encoders and measure similarities between two objects 
by learning a metric. \newcite{wu-etal-2019-open} designed RSN, utilizing neural siamese networks to determine whether two sentences express the same relation. In conventional bootstrapping systems, patterns are used to select new instances. Since neural networks bring better generalization than patterns, we use RSN to select high-confidence new instances by comparing candidates with existing ones.

Experiment results show that Neural Snowball achieves significant improvements on learning novel relations in few-shot scenarios. Further experiments demonstrate the efficiency of Relational Siamese Networks and the snowball process, proving that they have the ability to select high-quality instances and extract new relation facts. 


To conclude, our main contributions are threefold:
\begin{itemize} 
\setlength{\itemsep}{2pt}
\setlength{\parsep}{2pt}
\setlength{\parskip}{2pt}
\item We propose Neural Snowball, a novel approach to better train neural relation classifiers with only a handful of instances for new relations, by iteratively accumulating novel instances and facts from unlabeled data with prior knowledge of existing relations.
\item For better selecting new supporting instances for new relations, we investigate Relational Siamese Networks (RSN) to measure relational similarities between candidate instances and existing ones.
\item Experiment results and further analysis show the efficiency and robustness of our models. 
\end{itemize}

\section{Related Work}

\paragraph{Supervised RE} Early work for fully-supervised RE uses kernel methods \cite{zelenko2003kernel} and embedding methods \cite{gormley2015improved} to leverage syntactic information to predict relations. Recently, neural models like RNN and CNN have been proposed to extract better features from word sequences \cite{socher2012semantic,zeng2014relation}. 
Besides, dependency parsing trees 
have also been proved to be efficient in RE \cite{xu2015classifying,liu2015dependency}. 

\paragraph{Distant Supervision} Supervised RE methods rely on hand-labeled corpora, which usually cover only a limited number of relations and instances. 
\newcite{mintz2009distant} propose distant supervision to automatically generate relation labels by aligning entities between corpora and KGs. To alleviate wrong labeling, \newcite{riedel2010modeling} and \newcite{hoffmann2011knowledge} model distant supervision as a multi-instance multi-label task.

\paragraph{RE for New Relations} Bootstrapping RE can fast adapt to new relations with a small set of seed facts or sentences. \newcite{brin1998extracting} first proposes to extract relation facts by iterative pattern expansion from web. \newcite{agichtein2000snowball} propose Snowball to improve such iterative mechanism with better pattern extraction and evaluation methods.
Based on that, \newcite{zhu2009statsnowball} adopt statistical methods for better pattern selection. \newcite{batista2015semi} use word embeddings to further improve Snowball. Many similar bootstrapping ideas have been widely explored for RE \cite{pantel2006espresso,rozenfeld2008self,nakashole2011scalable}. 

Compared to distant supervision, bootstrapping expands relation facts iteratively, leading to higher precision. Moreover, distant supervision is still limited to predefined relations, yet bootstrapping is scalable for open-ended relation growth. Many other semi-supervised methods can also be adopted for RE \cite{rosenberg2005semi,french2017self,lin2019learning}, yet they still require sufficient annotations and mainly aim at classifying predefined relations rather than discovering new ones. Thus, we do not further discuss these methods. 


Inspired by the fact that people can grasp new knowledge with few samples, few-shot learning to solve data deficiency appeals to researchers. The key point of few-shot learning is to transfer task-agnostic information from existing data to new tasks \cite{bengio2012deep}. \newcite{vinyals2016matching}, \newcite{snell2017prototypical} and \newcite{zhang2018deep} explore learning a distance distribution to classify new classes in a nearest-neighbour-style strategy. \newcite{ravi2016optimization}, \newcite{munkhdalai2017meta} and \newcite{finn2017model} propose meta-learning to understand how to fast optimize models with few samples. \newcite{qiao2018few} propose learning to predict parameters for classifiers of new tasks. Existing few-shot learning models mainly focus on vision tasks. For exploiting it on text, \newcite{han2018fewrel} release FewRel, a large-scale few-shot RE dataset.

\paragraph{OpenRE} Both bootstrapping and few-shot learning handle new tasks with minimal human participation. 
Open relation extraction (OpenRE), on the other hand, aims at extracting relations from text without predefined types. 
One kind of OpenRE systems focuses on finding relation mentions 
\cite{banko2007open}, while others exploit to form relation types automatically by clustering semantic patterns \cite{shinyama2006preemptive,yao2011structured,elsahar2017unsupervised}.
It is a different and challengeable view on RE compared to conventional methods and remains to be explored.

\paragraph{Siamese Networks} Siamese networks measure similarities between two objects with dual encoders and trainable distance functions \cite{bromley1994signature}. 
They are exploited for one/few-shot learning \cite{koch2015siamese} and measuring text similarities \cite{mueller2016siamese}. 
\newcite{wu-etal-2019-open} propose Relational Siamese Networks (RSN) to learn a relational metric between given instances. Here we use RSN to select high-confidence instances by comparing candidates with existing ones.

\begin{figure*}[t]
\centering
\includegraphics[width=0.98\textwidth]{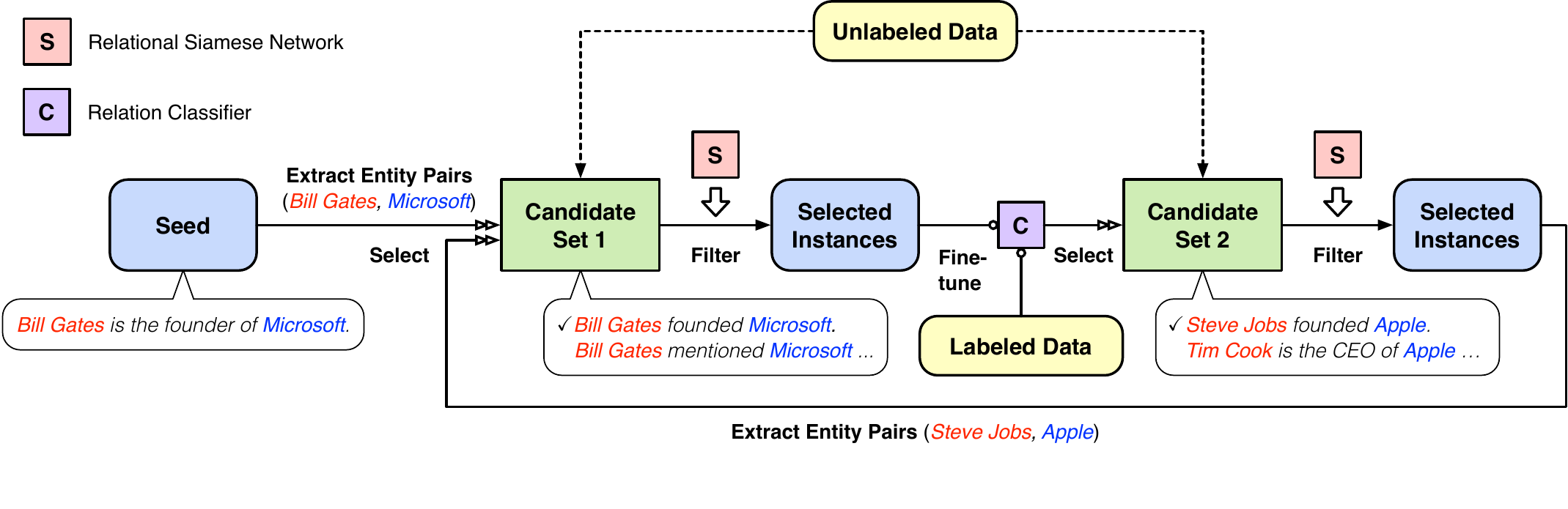}
\caption{The framework of Neural Snowball with examples of the relation \emph{founder}. Candidate set 1 ($\mathcal{C}_1$) contains all instances that have the same entity pairs as extracted. Candidate set 2 ($\mathcal{C}_2$) consists of high-confidence instances selected by the relation classifier. Instances in both candidate sets are filtered by RSN and then added to the selected instance set $\mathcal{S}_r$ of the relation $r$.
}
\label{fig:structure}
\end{figure*}

\section{Methodology}

In this section, we will introduce Neural Snowball, starting with notations and definitions.

\subsection{Terminology and Problem Definition}
\label{term}
Given an instance $x$ containing a word sequence $\{w_1,w_2,...,w_l\}$ with tagged entities $e_h$ and $e_t$, RE aims at predicting the relation label $r$ between $e_h$ and $e_t$. 
\textbf{Relation mentions} are instances expressing given relations.
\textbf{Entity pair mentions} are instances with given entity pairs.
\textbf{Relation facts} are triplets $(e_h,r,e_t)$ indicating there is a relation $r$ between $e_h$ and $e_t$.
$x^{r}$ indicates $x$ is a relation mention of the relation $r$.


Since we emphasize learning to extract a new relation in a real-world scenario, we adopt a different problem setting from existing supervised RE or few-shot RE. Given a large-scale labeled dataset for existing relations and a small set of instances for the new relation, our goal is to extract instances of the new relation from a query set containing instances of existing relations, the new relation and unseen relations.

Inputs of this task contain 
a large-scale labeled corpus $\mathcal{S}_N=\{x_j^{r_i}|r_i\in \mathcal{R}_N\}$ 
where $\mathcal{R}_N$ is a predefined relation set, 
an unlabeled corpus $\mathcal{T}$ 
and a seed set $\mathcal{S}_r$ with $k$ instances for the new relation $r$. 
We firstly pre-train the neural modules on $\mathcal{S}_N$. 
Then for the new relation $r$, we train a binary classifier $g$. 
To be more specific, given an instance $x$, $g(x)$ outputs the probability that $x$ expresses the relation $r$. 
During the test phase, the classifier $g$ performs classification on a query set $\mathcal{Q}$ containing instances expressing predefined relations in $\mathcal{R}_N$, instances with the new relation $r$ and some instances of other unseen relations, which is a simulation of the real-world scenario.

\subsection{Neural Snowball Process}
\label{sec:process}

Neural Snowball gathers reliable instances for a new relation $r$ iteratively with a small seed set $\mathcal{S}_r$ as the input. 
In each iteration, $\mathcal{S}_r$ will be extended with selected unlabeled instances, and the new $\mathcal{S}_r$ becomes the input of the next iteration. Figure \ref{fig:structure} illustrates the framework of Neural Snowball. When a new relation arrives with its initial instances, Neural Snowball shall process as follows,

\paragraph{Input} The seed instance set $\mathcal{S}_r$ for the relation $r$.

\paragraph{Phase 1} Structure the entity pair set,
\begin{equation}
    \mathcal{E}=\{(e_h,e_t)|\texttt{Ent}(x)=(e_h,e_t), x\in \mathcal{S}_r\},
\end{equation} 
where $\texttt{Ent}(x)$ means the entity pair of the instance $x$. Then, we get the candidate set $\mathcal{C}_1$ from the corpus $\mathcal{T}$ with
\begin{equation}
\mathcal{C}_1=\{x|\texttt{Ent}(x)\in \mathcal{E},x\in \mathcal{T}\}.
\end{equation}

Since those instances in $\mathcal{C}_1$ share same entity pairs with those in $\mathcal{S}_r$, we believe that they are likely to express the relation $r$. Yet to further alleviate false positive instances, for each $x$ in $\mathcal{C}_1$, we pair it with all instances $x'\in \mathcal{S}_r$ that share the same entity pair with $x$, and use the Relational Siamese Network (RSN) to get similarity scores. Averaging those scores we will get a confidence score of $x$, noted as $\texttt{score}_1(x)$.

Then, we sort instances in $\mathcal{C}_1$ in decreasing order of confidence scores and pick the top-$K_1$ instances as new ones added to $\mathcal{S}_r$. Since there exists the circumstance that less than $K_1$ instances really belong to the relation, we add an external condition that instances with confidence scores less than a threshold $\alpha$ will be excluded.

After all these steps, we have acquired new instances for the relation $r$ with high confidence. With the expanded instance set $\mathcal{S}_r$, we can fine-tune the relation classifier $g$ as described later, for the classifier is needed in the next step.

\paragraph{Phase 2} In the last phase, we expand $\mathcal{S}_r$, yet the entity pair set remains the same. So in this phase, 
our goal is to discover instances with new entity pairs for the relation $r$. 
We construct the candidate set for this phase by using the relation classifier $g$,
\begin{equation}
\mathcal{C}_2=\{x|g(x)>\theta,x\in \mathcal{T} \},
\end{equation}
where $\theta$ is a confidence threshold. 
Then each candidate instance $x$ is paired with each $x'$ in $\mathcal{S}_r$ as input of RSN, and the confidence score $\texttt{score}_2(x)$ is the mean of all the similairy scores of those pairs. 
Instances having top-$K_2$ scores and with $\texttt{score}_2$ larger than threshold $\beta$ are added to $\mathcal{S}_r$.

After one iteration of the process, we go back to phase 1, and another round starts. As the system runs, the instance set $\mathcal{S}_r$ grows bigger and the performance of the classifier increases until it reaches the peak. Best choices of the number of iterations and parameters mentioned above are discussed in the experiment section.

\subsection{Neural Modules}
\label{sec:modules}


Neural Snowball contains two key components: 
(1) the \textbf{Relational Siamese Network (RSN)}, which aims at selecting high-quality instances from unlabeled data by measuring similarities between candidate instances and existing ones, and (2) the \textbf{Relation Classifier}, which classifies whether an instance belongs to the new relation.

\begin{figure}[t]
\centering
\includegraphics[width=0.45\textwidth]{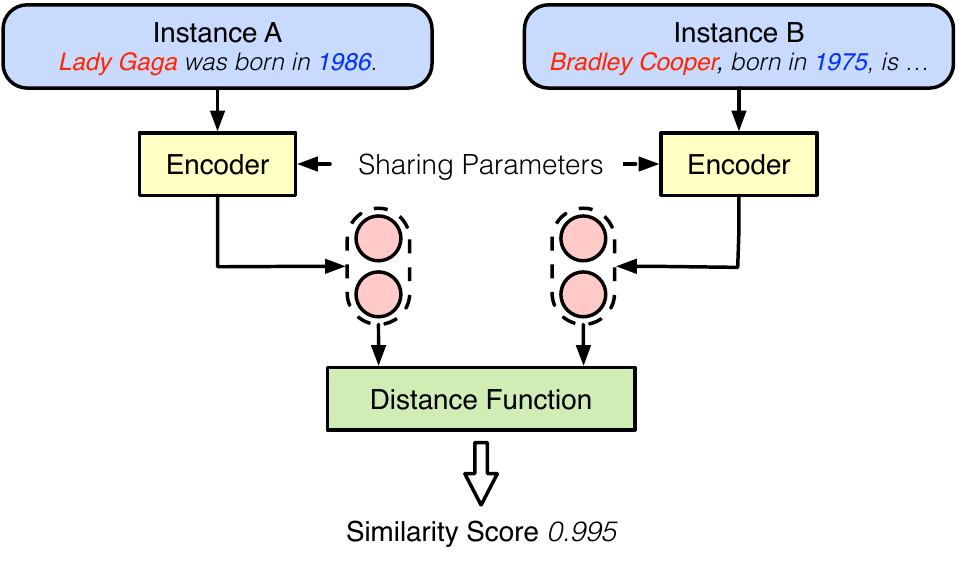}
\caption{The architecture of Relational Siamese Network (RSN). The encoders produce the representations of instances, and then RSN measures the similarity between them with certain distance function.}
\label{fig:siamese}
\end{figure}

\paragraph {Relational Siamese Network (RSN) $s(x,y)$} It takes two instances as input and outputs a value between 0 and 1 indicating the probability that those two instances share the same relation type. Figure \ref{fig:siamese} shows the structure of our proposed Relational Siamese Network, which consists of two encoders $f_s$ sharing parameters and a distant function. With instances as input, those encoders output the representation vectors for them. Then we compute the similarity score between the two instances with the following formula,
\begin{equation}
    s(x,y)=\sigma\big (\mathbf{w}_s^T (f_s(x) - f_s(y))^2 +b_s \big),
    \label{equ:siamese}
\end{equation}
where the square notation refers to squaring each dimension of the vector instead of the dot production of the vector, and $\sigma(\cdot)$ refers to sigmoid function.
This distance function can be considered as a weighted L2 distance with trainable weights $\mathbf{w}_s$ and bias $b_s$. A higher score indicates a higher possibility that the two sentences express the same relation ($\mathbf{w}_s$ will be negative to make this possible).

\paragraph{Relation Classifier $g(x)$} The classifier is composed of a neural encoder $f$, which transfers the raw instance $x$ into a real-valued vector, and a linear layer with parameters $\mathbf{w}$ and $b$ to get the probability that the input instance belongs to a relation $r$. It can be described by the following expression,
\begin{equation}
    g(x) = \sigma \big (\mathbf{w}^T f(x) + b\big ),
\end{equation}
where $g(x)$ is the output probability and $\sigma(\cdot)$ is sigmoid function to constrain the output between $0$ and $1$. Note that it is a binary classifier so $g(x)$ is just one real value, instead of a vector in the N-way classification scenario.

The reason to set it as a binary classifier instead of training an N-way classifier and utilizing softmax to constrain the outputs is that real-world relation extraction systems need to deal with negative samples, which express unknown relations and occupy a large proportion in corpora. These negative representations are not clusterable and considering them as ``one class'' is inappropriate. Another reason is that by using binary classifiers, we can handle the emergence of new relations by adding new classifiers, while the N-way classifier has to be retrained and data unbalance may lead to worse results for both new and existing relations.

With N binary classifiers, we can do N-way classification by comparing the output of each classifier, and the one with the highest probability wins. When no output exceeds a certain threshold, the sentence will be regarded as ``negative'', which means it does not express any of the existing relations.

\paragraph{Pre-training and Fine-tuning} 

To measure instance similarities on a new relation and to fast adapt the classifier to a new task, we need to pre-train the two neural modules. With the existing labeled dataset $\mathcal{S}_N$, we can perform a supervised N-way classification to pre-train the hidden representations of the classifier. As for RSN, we randomly sample instance pairs with the same or different relations from $\mathcal{S}_N$ and train the model with a cross entropy loss. 



When given a new relation $r$ with its $\mathcal{S}_r$, the parameters for the whole RSN and the encoder of the relation classifier are fixed, since they have already learned to extract generic features during pre-training. 
Further fine-tuning those parts with a small number of data might bring noise and bias to the distribution of the parameters.

Then we optimize the linear layer parameters $\mathbf{w}$ and $b$ in the classifier by sampling minibatches from $\mathcal{S}_r$ as positive samples and from $\mathcal{S}_N$ as negative samples. Denoting the positive batch as $\mathcal{S}_b$ and the negative batch as $\mathcal{T}_b$, the loss is as follows,

\begin{equation}
\begin{split}
\mathcal{L}_{\mathcal{S}_b,\mathcal{T}_b}(g_{\mathbf{w},b})&= \sum \limits_{x\in \mathcal{S}_{b}}{\log g_{\mathbf{w},b}(x)}\\ &+\mu\sum \limits_{x\in \mathcal{T}_{b}}{\log (1-g_{\mathbf{w},b}(x))}
\end{split}
\end{equation}

where $\mu$ is a coefficient of the negative sampling loss. Though for each batch we can sample positive and negative set with the same size, the actual numbers of positive instances and negative instances for the new relation differ a lot (a few versus thousands). So it is necessary to give the negative part of loss a smaller weight.

With the sampling strategy and loss function, we can do gradient-based optimization on parameters $\mathbf{w}$ and $b$. Here we choose Adam \cite{kingma2014adam} as our optimizer. The hyperparameters include the number of training epochs $e$, batch size $bs$, learning rate $\lambda$ and coefficient of negative sampling loss $\mu$. Algorithm \ref{algorithm} describes the process.

The fine-tuning process is used as one of our baselines. We also adopt this algorithm in each step of Neural Snowball after gathering new instances in $\mathcal{S}_r$. Though it is a simple way to acquire $\mathbf{w}$ and $b$, it is better than metric-based few-shot algorithms for that it is more adaptive to new relations while metric-based models usually fix all the parameters during few-shot, and it is more scalable to a large number of training instances. Negative sampling also enables the model to improve the precision of extracting new relation.

\subsection{Neural Encoders}
\label{encoders}

As mentioned above, encoders are parts of our RSN and classifiers and aim at extracting abstract and generic features from raw sentences and tagged entities. 
In this paper, we adopt two encoders: CNN \cite{nguyen2015relation} and BERT \cite{devlin2018bert}. 

\paragraph{CNN} We follow the model structure in \newcite{nguyen2015relation} for our CNN encoder. The model takes word embeddings and position embeddings \cite{zeng2014relation} as input. The embedding sequence is then fed into a one-dim convolutional neural network to extract features. Then those features are max-pooled to get one real-valued vector as the instance representation.

\paragraph{BERT} \newcite{devlin2018bert} propose a novel language model named BERT, which stands for Bidirectional Encoder Representations from Transformers, 
and has obtained new state-of-the-arts on several NLP tasks, far beyond existing CNN or RNN models. BERT takes tokens of the sentence as input and after several attention layers outputs hidden features for each token. To fit the RE task, we add special marks at the beginning of the sequence and before and after the entities. Note that marks at the beginning, around the head entities and tail entities are different. Then, we take the hidden features of the first token as the sentence representation. 

\begin{algorithm}[t]
\small
\caption{Fine-tuning the Classifier}
\label{algorithm}
\KwIn{New instance set $\mathcal{S}_r$, historical relation dataset $\mathcal{S}_N$}
\KwResult{Optimized $\mathbf{w}$ and $b$}
Randomly initialize $\mathbf{w}$ and $b$

\For{$i\gets1$ \KwTo $e$} {
    // Get a sequence of minibatches from $\mathcal{S}_r$

    $\mathcal{S}_{batch\_seq}\gets$batch\_seq($\mathcal{S}_r$,$bs$) 
    
    \For{$\mathcal{S}_{b}\in \mathcal{S}_{batch\_seq}$}{
        // Sample the negative batch
        
        $\mathcal{T}_{b}\gets$sample($\mathcal{S}_N$,$bs$)
        
        Update $\mathbf{w}$ and $b$ w.r.t. $\mathcal{L}_{\mathcal{S}_b,\mathcal{T}_b}(g_{\mathbf{w},b})$ 
        
        ~~~~with learning rate $\lambda$
    }
}

\end{algorithm}
 
 \begin{table*}[t]
    \centering
    \renewcommand\arraystretch{1.2}
        \begin{tabular}{c|ccccccccc}
        \toprule
        \multirow{2}{*}{Model} & \multicolumn{3}{c}{5 Seed Instances} & \multicolumn{3}{c}{10 Seed Instances} & \multicolumn{3}{c}{15 Seed Instances}\\
        & P & R & F1 & P & R & F1 & P & R & F1\\
        \midrule
BREDS & $33.71$ & $11.89$ & $17.58$ & $28.29$ & $17.02$ &$21.25$ & $25.24$ &$17.96$  & $20.99$ \\
\midrule
Fine-tuning (CNN)                & $46.90$ & $9.08$  & $15.22$ & $47.58$ & $38.36$ & $42.48$ & $74.70$ & $48.03$ & $58.46$ \\
 Relational Siamese Network (CNN) & $45.00$ & $31.37$ & $36.96$ & $46.42$ & $30.68$ & $36.94$ & $49.32$ & $30.46$ & $37.66$\\
 Distant Supervision (CNN)                       & $44.99$ & $31.06$ & $36.75$ & $42.48$ & $48.64$ & $45.35$ & $43.70$ & $54.76$ & $48.60$\\
 \textbf{Neural Snowball (CNN)} & $48.07$ & $36.21$ & $41.30$ & $47.28$ & $51.49$ & $49.30$ & $68.25$ & $58.90$ & $63.23$\\
        \midrule
         
Fine-tuning (BERT)               & $50.85$ & $16.66$          & $25.10$ & $59.87$ & $55.19$          & $57.43$ & $\textbf{81.60}$ & $58.92$ & $68.43$\\
 Relational Siamese Network (BERT) & $39.07$ & $\textbf{51.39}$ & $44.47$     & $42.42$ & $54.93$      & $47.87$ & $44.10$ & $52.73$ & $48.03$ \\
Distant Supervision (BERT)                       & $38.06$ & $51.18$          & $43.66$ & $38.45$ & $\textbf{76.12}$ & $51.09$ & $35.48$ & $\textbf{80.33}$ & $49.22$ \\
 \textbf{Neural Snowball (BERT)} & $\textbf{56.87}$ & $40.43$ &$\textbf{47.26}$ &$\textbf{60.50}$ &$62.20$ &$\textbf{61.34}$ & $78.13$& $66.87$& $\textbf{72.06}$\\
  \bottomrule
    \end{tabular}
    \caption{Experiment results on our few-shot relation learning settings with different size of seed sets. Here P refers to precision, R refers to recall and F1 refers to F1-measure score.}
    \label{table:exp}
    \end{table*}

\section{Experiments}

In this section, we will show that the relation classifiers trained with our Neural Snowball mechanism achieve significant improvements compared to baselines in our few-shot relation learning settings. 
We also carry out two quantitative evaluations to further prove the effectiveness of Relational Siamese Networks and the snowball process.

\subsection{Datasets and Evaluation Settings}

Our experiment setting requires a dataset with precise human annotations, large amount of data and also it needs to be easy to perform distant supervision on. For now the only qualified dataset is FewRel \cite{han2018fewrel}. It contains 100 relations and 70,000 instances from Wikipedia. The dataset is divided into three subsets: training set (64 relations), validation set (16 relations) and test set (20 relations). We also dump an unlabeled corpus from Wikipedia with tagged entities, including 899,996 instances and 464,218 entity pairs, which is used for the snowball process.


Our main experiment follows the setting in previous sections. First we further split the training set into training set A and B. We use the training set A as $\mathcal{S}_N$, and for each step of evaluation, we sample one relation as the new relation $r$ and $k$ instances of it as $\mathcal{S}_r$ from val/test set, and sample a query set $\mathcal{Q}$ from both training set B and val/test set. Then the models classify all the query instances in a binary manner, judging whether each instance mentions the new relation $r$. Note that the sampled query set includes $N$ relations with sufficient training data, one relation $r$ with few instances and many other unseen relations. It is a very challengeable setting and closer to the real-world applications compared to N-way K-shot few-shot (sampling N classes and classifying inside the N classes), since corpora in the real world are not limited to certain relation numbers or types.




\subsection{Parameter Settings}

We tune our hyperparameters on the validation set. For parameters of the encoders, we follow \cite{han2018fewrel} for CNN and \cite{devlin2018bert} for BERT. 
For the fine-tuning, after grid searching, we adopt training epochs $e=50$, batch size $bs=10$, learning rate $\lambda=0.05$ and negative loss coefficient $\mu=0.2$. 
BERT fine-tuning shares the same parameters except for $\lambda=0.01$ and $\mu=0.5$. 

For the Neural Snowball process, we also determine our parameters by grid searching. We set $K_1$ and $K_2$, the numbers of added instances for each stage, as $5$, and the thresholds of RSN for each stage, $\alpha$ and $\beta$, as $0.5$. We adopt $0.9$ for the classifier threshold $\theta$.

All the models evaluated in our experiments output a probability of being the mention of the new relation for each query instance, and to get the predicting results we need to set a confidence threshold. For fine-tuning and Neural Snowball we set the threshold as $0.5$, and $0.7$ for the Relational Siamese Network. 

\subsection{Few-Shot Relation Learning}

Table \ref{table:exp} shows the experiment results on our few-shot relation learning tasks. We evaluate five model architectures: \textbf{BREDS} \cite{batista2015semi} is an advanced version of the original snowball \cite{agichtein2000snowball}, which uses word embeddings for pattern selection; \textbf{Fine-tuning} stands for directly using Algorithm \ref{algorithm} with few-shot instances to train the new classifier;
\textbf{Relational Siamese Network (RSN)} refers to computing similarity scores between the query instance and each instance in $\mathcal{S}_r$, and averaging them as the probability of the query one expressing the new relation; \textbf{Distant Supervision} refers to taking all instances sharing entity pairs with given seeds into the training set and using Algorithm \ref{algorithm}; \textbf{Neural Snowball} is our proposed method. 
We do not evaluate other semi-supervised and few-shot RE models for the reason that they do not suit our few-shot new relation learning settings.



From Table \ref{table:exp} we can identify that (1) our Neural Snowball achieves the best results in both settings and with both encoders. 
(2) While fine-tuning, distant supervision and Neural Snowball improve with the increase of seed numbers, BREDS and RSN have little promotion. 

By further comparison between Neural Snowball and other baselines, we notice that our model largely promotes the recall values while maintaining the high precision values. It indicates that Neural Snowball not only gathers new training instances with high quality, but also successfully extracts new relation facts and patterns to widen the coverage of instances for the new relation.

\subsection{Analysis on Relational Siamese Network}

\begin{table}[!hbtp]
    \centering
    \renewcommand\arraystretch{1.2}
    \begin{tabular}{c|cccc}
        \toprule
        Relation Set & P@5 & P@10 & P@20 & P@50\\
        \midrule
        Train & 83.60 & 80.66 & 76.03 & 61.98\\
        Test & 82.15 & 78.64 & 72.57 & 55.10\\
        \bottomrule
    \end{tabular}
    \caption{Precisions at top-N instances scored by RSN (CNN) in the 5-seed setting. ``Train'' and ``Test'' represent results on relations in the training and test sets.}
    \label{table:pn}
\end{table} 

To examine the quality of instances selected by RSN, we randomly sample one relation and 5 instances of it and use the rest data as query instances. We use the method mentioned before to calculate a score for each query instance, then we calculate precisions at top-$N$ instances (P@$N$).

We can see that RSN achieves a precision of $82.15\%$ at top-5 instances on the test set. It is relative high considering RSN is only given a small number of instances and it even have not seen the relation before. Also note that though RSN is only trained with relations of the training set, the performance on relations in the test set has only a narrow gap, further proving the effectiveness of RSN. 

\begin{figure}[t!]
\centering
\begin{subfigure}{0.45\textwidth}
\includegraphics[width=\textwidth]{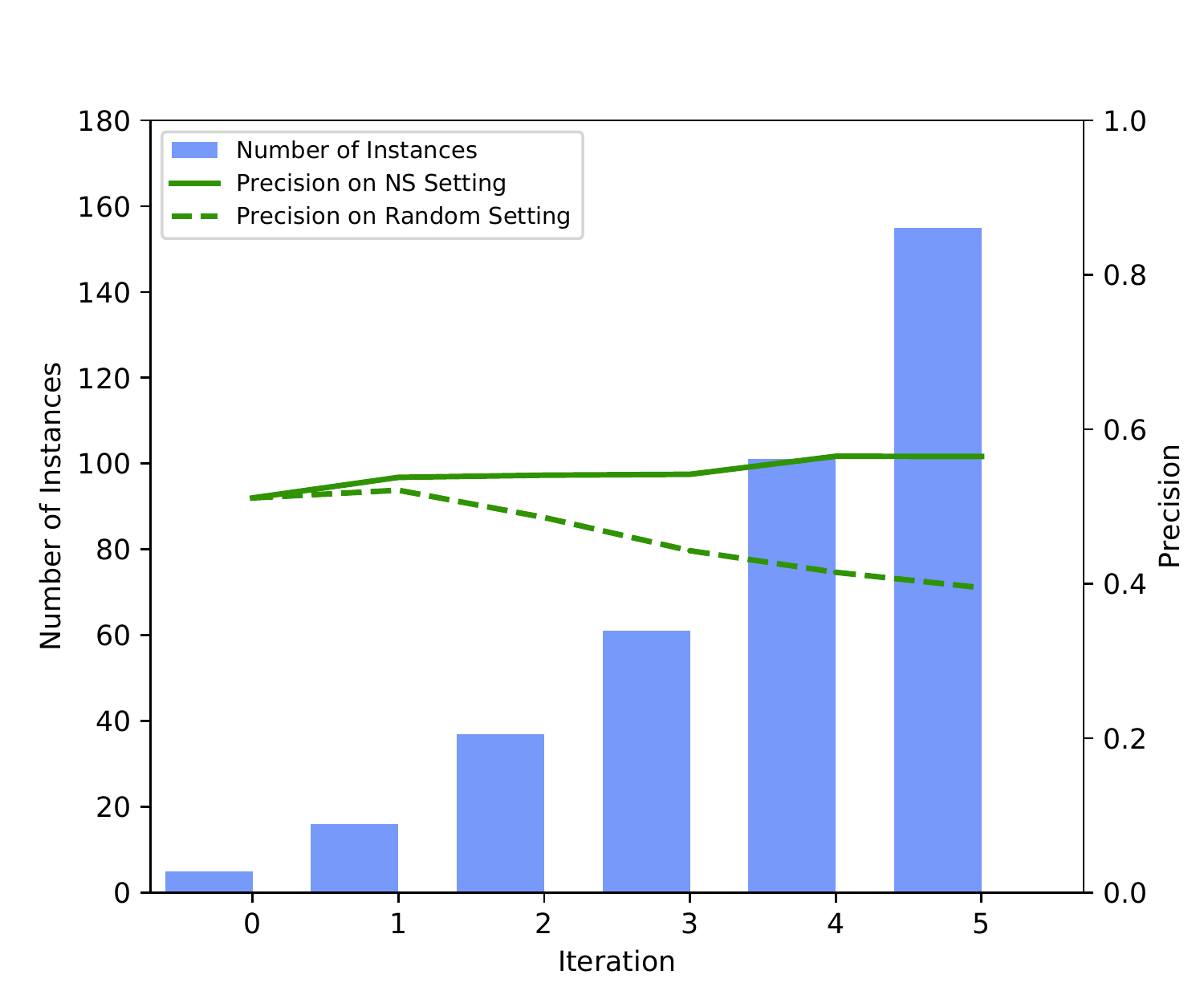}
\end{subfigure}
\begin{subfigure}{0.45\textwidth}
\includegraphics[width=\textwidth]{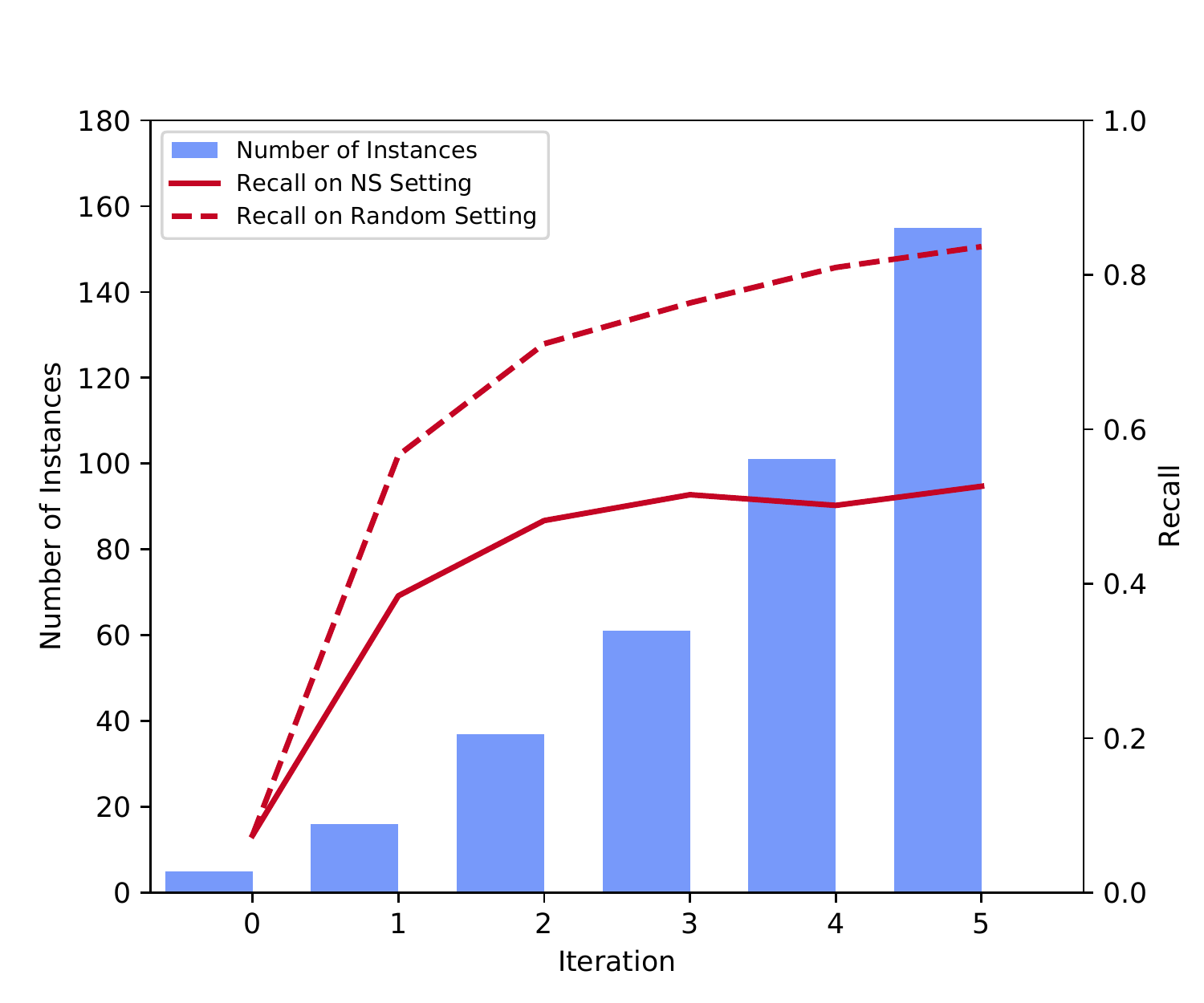} 
\end{subfigure}
\caption{Evaluation results on each iteration of Neural Snowball. Blue bars are numbers of instances added. Solid lines represent performance on the NS setting, and dotted lines represent the random setting.}
\label{fig:precrec}
\end{figure}

\subsection{Analysis on Neural Snowball Process}

To further analyze the iterative process of Neural Snowball (NS), we present a quantitative evaluation on the numbers of newly-gathered instances as well as the classifier performance on relation \emph{chairperson} with the 5-seed-instance setting. 
Note that it is a randomly-picked relation and other relations have shown similar trends. Due to the space limit, we only take the relation \emph{chairperson} as an example.

Figure \ref{fig:precrec} demonstrates the development of evaluation results as the iteration grows. 
Here we adopt two settings: \textbf{NS setting} refers to fine-tuning the classifier with instances selected by Neural Snowball, and \textbf{random setting} refers to fine-tuning on randomly-picked instances of relation \emph{chairperson} with the same amount of NS, under the premise of knowing all the instances of the relation. Note that random setting is an ideal case since it reflects the real distribution of data for the new relation and the overall performance of the random setting serves as an upper bound. 



From the results of random setting, we see that the binary classifier obtains higher recall and performs a little lower in precision when trained on larger randomly-distributed data.
This can be explained that more data brings more patterns in representations, improving the completeness of extracting while sacrificing a little in quality.

Then by comparing the results between the two settings, we get two observations: 
(1) As the number of iterations and amount of instances grow, the classifier fine-tuned on NS setting maintains higher precision than the one fine-tuned on random setting, which proves that RSN succeeds in extracting high-confidence instances and brings in high-quality patterns. 
(2) The recall rate of NS grows less than expected, indicating that RSN 
might overfit existing patterns. 
To maintain high precision of the model, Neural Snowball stucks in the ``comfort zone''of existing high-quality patterns and fails to jump out of the zone to discover patterns with more diversity. 
We plan to further investigate it in future.




\section{Conclusion and Future Work}

In this paper, we propose Neural Snowball, a novel approach that learns to classify a new relation with only a small number of instances. We use Relational Siamese Networks (RSN), which are pre-trained on historical relations to iteratively select reliable instances for the new relation from unlabeled corpora. 
Evaluations on a large-scale relation extraction dataset demonstrate that 
Neural Snowball brings significant improvement in performance of extracting new relations with few instances. 
Further analysis proves the effectiveness of RSN and the snowball process. In the future, we will further explore the following directions: 

(1) The deficiency of our current model is that it mainly extracts patterns semantically close to the given instances, which limits the increase in recall. In the future, we will explore how to jump out of the ``comfort zone'' and discover instances with more diversity.

(2) For now, RSN is fixed during new relation learning and shares the same parameters across relations.
This can be ameliorated by an adaptive RSN that can be further optimized given new relations and new instances. 
We will investigate into it and further improve the efficiency of RSN.

\section{Acknowledgments}

This work is supported by the National Natural Science Foundation of China (NSFC No. 61572273, 61661146007, 61772302) and the research fund of Tsinghua University - Tencent Joint Laboratory for Internet Innovation Technology. Han and Gao are supported by 2018 and
2019 Tencent Rhino-Bird Elite Training Program
respectively. Gao is also supported by Tsinghua
University Initiative Scientific Research Program.

\bibliography{aaai20}

\begin{thebibliography}{}

\bibitem[\protect\citeauthoryear{Agichtein and
  Gravano}{2000}]{agichtein2000snowball}
Agichtein, E., and Gravano, L.
\newblock 2000.
\newblock Snowball: Extracting relations from large plain-text collections.
\newblock In {\em Proceedings of JCDL},  85--94.

\bibitem[\protect\citeauthoryear{Banko \bgroup et al\mbox.\egroup
  }{2007}]{banko2007open}
Banko, M.; Cafarella, M.~J.; Soderland, S.; Broadhead, M.; and Etzioni, O.
\newblock 2007.
\newblock Open information extraction from the web.
\newblock In {\em Proceedings of IJCAI},  2670--2676.

\bibitem[\protect\citeauthoryear{Batista, Martins, and
  Silva}{2015}]{batista2015semi}
Batista, D.~S.; Martins, B.; and Silva, M.~J.
\newblock 2015.
\newblock Semi-supervised bootstrapping of relationship extractors with
  distributional semantics.
\newblock In {\em Proceedings of EMNLP},  499--504.

\bibitem[\protect\citeauthoryear{Bengio}{2012}]{bengio2012deep}
Bengio, Y.
\newblock 2012.
\newblock Deep learning of representations for unsupervised and transfer
  learning.
\newblock In {\em Proceedings of the Workshop on Unsupervised and Transfer
  Learning of ICML},  17--36.

\bibitem[\protect\citeauthoryear{Bollacker \bgroup et al\mbox.\egroup
  }{2008}]{bollacker2008freebase}
Bollacker, K.; Evans, C.; Paritosh, P.; Sturge, T.; and Taylor, J.
\newblock 2008.
\newblock Freebase: a collaboratively created graph database for structuring
  human knowledge.
\newblock In {\em Proceedings of SIGMOD},  1247--1250.

\bibitem[\protect\citeauthoryear{Brin}{1998}]{brin1998extracting}
Brin, S.
\newblock 1998.
\newblock Extracting patterns and relations from the world wide web.
\newblock In {\em Proceedings of International Workshop on The World Wide Web
  and Databases},  172--183.

\bibitem[\protect\citeauthoryear{Bromley \bgroup et al\mbox.\egroup
  }{1994}]{bromley1994signature}
Bromley, J.; Guyon, I.; LeCun, Y.; S{\"a}ckinger, E.; and Shah, R.
\newblock 1994.
\newblock Signature verification using a" siamese" time delay neural network.
\newblock In {\em Proceedings of NIPS},  737--744.

\bibitem[\protect\citeauthoryear{Devlin \bgroup et al\mbox.\egroup
  }{2019}]{devlin2018bert}
Devlin, J.; Chang, M.-W.; Lee, K.; and Toutanova, K.
\newblock 2019.
\newblock {BERT}: Pre-training of deep bidirectional transformers for language
  understanding.
\newblock In {\em Proceedings of NAACL-HLT},  4171--4186.

\bibitem[\protect\citeauthoryear{ElSahar \bgroup et al\mbox.\egroup
  }{2017}]{elsahar2017unsupervised}
ElSahar, H.; Demidova, E.; Gottschalk, S.; Gravier, C.; and Laforest, F.
\newblock 2017.
\newblock Unsupervised open relation extraction.
\newblock In {\em Proceedings of ESWC},  12--16.

\bibitem[\protect\citeauthoryear{Finn, Abbeel, and
  Levine}{2017}]{finn2017model}
Finn, C.; Abbeel, P.; and Levine, S.
\newblock 2017.
\newblock Model-agnostic meta-learning for fast adaptation of deep networks.
\newblock In {\em Proceedings of ICML},  1126--1135.

\bibitem[\protect\citeauthoryear{French, Mackiewicz, and
  Fisher}{2017}]{french2017self}
French, G.; Mackiewicz, M.; and Fisher, M.
\newblock 2017.
\newblock Self-ensembling for visual domain adaptation.
\newblock {\em arXiv preprint arXiv:1706.05208}.

\bibitem[\protect\citeauthoryear{Gormley, Yu, and
  Dredze}{2015}]{gormley2015improved}
Gormley, M.~R.; Yu, M.; and Dredze, M.
\newblock 2015.
\newblock Improved relation extraction with feature-rich compositional
  embedding models.
\newblock In {\em Proceedings of EMNLP},  1774--1784.

\bibitem[\protect\citeauthoryear{Han \bgroup et al\mbox.\egroup
  }{2018}]{han2018fewrel}
Han, X.; Zhu, H.; Yu, P.; Wang, Z.; Yao, Y.; Liu, Z.; and Sun, M.
\newblock 2018.
\newblock Fewrel: A large-scale supervised few-shot relation classification
  dataset with state-of-the-art evaluation.
\newblock In {\em Proceedings of EMNLP},  4803--4809.

\bibitem[\protect\citeauthoryear{Hoffmann \bgroup et al\mbox.\egroup
  }{2011}]{hoffmann2011knowledge}
Hoffmann, R.; Zhang, C.; Ling, X.; Zettlemoyer, L.; and Weld, D.~S.
\newblock 2011.
\newblock Knowledge-based weak supervision for information extraction of
  overlapping relations.
\newblock In {\em Proceedings of ACL-HLT},  541--550.

\bibitem[\protect\citeauthoryear{Kingma and Ba}{2015}]{kingma2014adam}
Kingma, D.~P., and Ba, J.
\newblock 2015.
\newblock Adam: A method for stochastic optimization.
\newblock In {\em Proceedings of ICLR}.

\bibitem[\protect\citeauthoryear{Koch, Zemel, and
  Salakhutdinov}{2015}]{koch2015siamese}
Koch, G.; Zemel, R.; and Salakhutdinov, R.
\newblock 2015.
\newblock Siamese neural networks for one-shot image recognition.
\newblock In {\em Proceedings of the Workshop of ICML}.

\bibitem[\protect\citeauthoryear{Lin \bgroup et al\mbox.\egroup
  }{2019}]{lin2019learning}
Lin, H.; Yan, J.; Qu, M.; and Ren, X.
\newblock 2019.
\newblock Learning dual retrieval module for semi-supervised relation
  extraction.
\newblock In {\em Proceedings of WWW},  1073--1083.

\bibitem[\protect\citeauthoryear{Liu \bgroup et al\mbox.\egroup
  }{2015}]{liu2015dependency}
Liu, Y.; Wei, F.; Li, S.; Ji, H.; Zhou, M.; and Houfeng, W.
\newblock 2015.
\newblock A dependency-based neural network for relation classification.
\newblock In {\em Proceedings of ACL-IJCNLP},  285--290.

\bibitem[\protect\citeauthoryear{Miller}{1995}]{miller1995wordnet}
Miller, G.~A.
\newblock 1995.
\newblock Wordnet: a lexical database for english.
\newblock {\em Communications of the ACM} 38(11):39--41.

\bibitem[\protect\citeauthoryear{Mintz \bgroup et al\mbox.\egroup
  }{2009}]{mintz2009distant}
Mintz, M.; Bills, S.; Snow, R.; and Jurafsky, D.
\newblock 2009.
\newblock Distant supervision for relation extraction without labeled data.
\newblock In {\em Proceedings of ACL-IJCNLP},  1003--1011.

\bibitem[\protect\citeauthoryear{Mueller and
  Thyagarajan}{2016}]{mueller2016siamese}
Mueller, J., and Thyagarajan, A.
\newblock 2016.
\newblock Siamese recurrent architectures for learning sentence similarity.
\newblock In {\em Proceedings of AAAI}.

\bibitem[\protect\citeauthoryear{Munkhdalai and Yu}{2017}]{munkhdalai2017meta}
Munkhdalai, T., and Yu, H.
\newblock 2017.
\newblock Meta networks.
\newblock In {\em Proceedings of ICML},  2554--2563.

\bibitem[\protect\citeauthoryear{Nakashole, Theobald, and
  Weikum}{2011}]{nakashole2011scalable}
Nakashole, N.; Theobald, M.; and Weikum, G.
\newblock 2011.
\newblock Scalable knowledge harvesting with high precision and high recall.
\newblock In {\em Proceedings of WSDM},  227--236.

\bibitem[\protect\citeauthoryear{Nguyen and
  Grishman}{2015}]{nguyen2015relation}
Nguyen, T.~H., and Grishman, R.
\newblock 2015.
\newblock Relation extraction: Perspective from convolutional neural networks.
\newblock In {\em Proceedings of the Workshop on Vector Space Modeling for
  NLP},  39--48.

\bibitem[\protect\citeauthoryear{Pantel and
  Pennacchiotti}{2006}]{pantel2006espresso}
Pantel, P., and Pennacchiotti, M.
\newblock 2006.
\newblock Espresso: Leveraging generic patterns for automatically harvesting
  semantic relations.
\newblock In {\em Proceedings of COLING/ACL},  113--120.

\bibitem[\protect\citeauthoryear{Qiao \bgroup et al\mbox.\egroup
  }{2018}]{qiao2018few}
Qiao, S.; Liu, C.; Shen, W.; and Yuille, A.~L.
\newblock 2018.
\newblock Few-shot image recognition by predicting parameters from activations.
\newblock In {\em Proceedings of CVPR},  7229--7238.

\bibitem[\protect\citeauthoryear{Ravi and
  Larochelle}{2017}]{ravi2016optimization}
Ravi, S., and Larochelle, H.
\newblock 2017.
\newblock Optimization as a model for few-shot learning.
\newblock In {\em Proceedings of ICLR}.

\bibitem[\protect\citeauthoryear{Riedel, Yao, and
  McCallum}{2010}]{riedel2010modeling}
Riedel, S.; Yao, L.; and McCallum, A.
\newblock 2010.
\newblock Modeling relations and their mentions without labeled text.
\newblock In {\em Proceedings of ECML-PKDD},  148--163.

\bibitem[\protect\citeauthoryear{Rosenberg, Hebert, and
  Schneiderman}{2005}]{rosenberg2005semi}
Rosenberg, C.; Hebert, M.; and Schneiderman, H.
\newblock 2005.
\newblock Semi-supervised self-training of object detection models.
\newblock In {\em Proceedings of WACV},  29--36.

\bibitem[\protect\citeauthoryear{Rozenfeld and
  Feldman}{2008}]{rozenfeld2008self}
Rozenfeld, B., and Feldman, R.
\newblock 2008.
\newblock Self-supervised relation extraction from the web.
\newblock {\em KAIS}  17--33.

\bibitem[\protect\citeauthoryear{Shinyama and
  Sekine}{2006}]{shinyama2006preemptive}
Shinyama, Y., and Sekine, S.
\newblock 2006.
\newblock Preemptive information extraction using unrestricted relation
  discovery.
\newblock In {\em Proceedings of NAACL-HLT},  304--311.

\bibitem[\protect\citeauthoryear{Snell, Swersky, and
  Zemel}{2017}]{snell2017prototypical}
Snell, J.; Swersky, K.; and Zemel, R.
\newblock 2017.
\newblock Prototypical networks for few-shot learning.
\newblock In {\em Proceedings of NIPS},  4077--4087.

\bibitem[\protect\citeauthoryear{Socher \bgroup et al\mbox.\egroup
  }{2012}]{socher2012semantic}
Socher, R.; Huval, B.; Manning, C.~D.; and Ng, A.~Y.
\newblock 2012.
\newblock Semantic compositionality through recursive matrix-vector spaces.
\newblock In {\em Proceedings of EMNLP-CoNLL},  1201--1211.

\bibitem[\protect\citeauthoryear{Vinyals \bgroup et al\mbox.\egroup
  }{2016}]{vinyals2016matching}
Vinyals, O.; Blundell, C.; Lillicrap, T.; Wierstra, D.; et~al.
\newblock 2016.
\newblock Matching networks for one shot learning.
\newblock In {\em Proceedings of NIPS},  3630--3638.

\bibitem[\protect\citeauthoryear{Vrande{\v{c}}i{\'c} and
  Kr{\"o}tzsch}{2014}]{vrandevcic2014wikidata}
Vrande{\v{c}}i{\'c}, D., and Kr{\"o}tzsch, M.
\newblock 2014.
\newblock Wikidata: a free collaborative knowledgebase.
\newblock {\em Communications of the ACM} 57(10):78--85.

\bibitem[\protect\citeauthoryear{Wu \bgroup et al\mbox.\egroup
  }{2019}]{wu-etal-2019-open}
Wu, R.; Yao, Y.; Han, X.; Xie, R.; Liu, Z.; Lin, F.; Lin, L.; and Sun, M.
\newblock 2019.
\newblock Open relation extraction: Relational knowledge transfer from
  supervised data to unsupervised data.
\newblock In {\em Proceedings of EMNLP-IJCNLP},  219--228.

\bibitem[\protect\citeauthoryear{Xu \bgroup et al\mbox.\egroup
  }{2015}]{xu2015classifying}
Xu, Y.; Mou, L.; Li, G.; Chen, Y.; Peng, H.; and Jin, Z.
\newblock 2015.
\newblock Classifying relations via long short term memory networks along
  shortest dependency paths.
\newblock In {\em Proceedings of EMNLP},  1785--1794.

\bibitem[\protect\citeauthoryear{Yao \bgroup et al\mbox.\egroup
  }{2011}]{yao2011structured}
Yao, L.; Haghighi, A.; Riedel, S.; and McCallum, A.
\newblock 2011.
\newblock Structured relation discovery using generative models.
\newblock In {\em Proceedings of EMNLP},  1456--1466.

\bibitem[\protect\citeauthoryear{Zelenko, Aone, and
  Richardella}{2003}]{zelenko2003kernel}
Zelenko, D.; Aone, C.; and Richardella, A.
\newblock 2003.
\newblock Kernel methods for relation extraction.
\newblock {\em JMLR}  1083--1106.

\bibitem[\protect\citeauthoryear{Zeng \bgroup et al\mbox.\egroup
  }{2014}]{zeng2014relation}
Zeng, D.; Liu, K.; Lai, S.; Zhou, G.; and Zhao, J.
\newblock 2014.
\newblock Relation classification via convolutional deep neural network.
\newblock In {\em Proceedings of COLING},  2335--2344.

\bibitem[\protect\citeauthoryear{Zhang \bgroup et al\mbox.\egroup
  }{2018}]{zhang2018deep}
Zhang, X.; Sung, F.; Qiang, Y.; Yang, Y.; and Hospedales, T.~M.
\newblock 2018.
\newblock Deep comparison: Relation columns for few-shot learning.
\newblock {\em arXiv preprint arXiv:1811.07100}.

\bibitem[\protect\citeauthoryear{Zhu \bgroup et al\mbox.\egroup
  }{2009}]{zhu2009statsnowball}
Zhu, J.; Nie, Z.; Liu, X.; Zhang, B.; and Wen, J.-R.
\newblock 2009.
\newblock Statsnowball: a statistical approach to extracting entity
  relationships.
\newblock In {\em Proceedings of WWW},  101--110.

\end{thebibliography}
\bibliographystyle{aaai}

\end{document}